\documentclass[conference]{IEEEtran}
\IEEEoverridecommandlockouts
\usepackage{cite}
\usepackage{amsmath,amssymb,amsfonts}
\usepackage{algorithmic}
\usepackage{graphicx}
\usepackage{textcomp}
\usepackage{xcolor}
\usepackage{colortbl}
\usepackage{textcomp}
\usepackage{tabularx}
\usepackage{booktabs}
\usepackage{url}
\usepackage{multirow}
\usepackage{hyperref}
\usepackage{adjustbox}
\usepackage{siunitx}

\def\BibTeX{{\rm B\kern-.05em{\sc i\kern-.025em b}\kern-.08em
    T\kern-.1667em\lower.7ex\hbox{E}\kern-.125emX}}
\begin{document}

\title{Context-Aware Pragmatic Metacognitive Prompting for Sarcasm Detection\\}

% \title{Addressing Linguistic and Domain-Specific Challenges in Sarcasm Detection through Context-Aware Pragmatic Metacognitive Prompting\\}

% \title{Addressing Linguistic and Domain-Specific Challenges in Sarcasm Detection through Context-Aware PMP\\}

% \title{Addressing Domain-Specific Challenges in Sarcasm Detection through Context-Aware PMP\\}

\author{\IEEEauthorblockN{1\textsuperscript{st} Michael Iskandardinata}
\IEEEauthorblockA{
\textit{Computer Science Department} \\
\textit{School of Computer Science} \\
\textit{Bina Nusantara University} \\
Jakarta, Indonesia \\
michael.iskandardinata@binus.ac.id}

\and

\IEEEauthorblockN{2\textsuperscript{nd} William Christian}
\IEEEauthorblockA{
\textit{Computer Science Department} \\
\textit{School of Computer Science} \\
\textit{Bina Nusantara University} \\
Jakarta, Indonesia \\
william.christian001@binus.ac.id}

\and

\IEEEauthorblockN{3\textsuperscript{rd} Derwin Suhartono}
\IEEEauthorblockA{
\textit{Computer Science Department} \\
\textit{School of Computer Science} \\
\textit{Bina Nusantara University} \\
Jakarta, Indonesia \\
dsuhartono@binus.edu}
}
\maketitle

\begin{abstract}
Detecting sarcasm remains a challenging task in the areas of Natural Language Processing (NLP) despite recent advances in neural network approaches. Currently, Pre-trained Language Models (PLMs) and Large Language Models (LLMs) are the preferred approach for sarcasm detection. However, the complexity of sarcastic text, combined with linguistic diversity and cultural variation across communities, has made the task more difficult even for PLMs and LLMs. Beyond that, those models also exhibit unreliable detection of words or tokens that require extra grounding for analysis. Building on a state-of-the-art prompting method in LLMs for sarcasm detection called Pragmatic Metacognitive Prompting (PMP), we introduce a retrieval-aware approach that incorporates retrieved contextual information for each target text. Our pipeline explores two complementary ways to provide context: adding non-parametric knowledge using web-based retrieval when the model lacks necessary background, and eliciting the model's own internal knowledge for a self-knowledge awareness strategy. We evaluated our approach with three datasets, such as Twitter Indonesia Sarcastic, SemEval-2018 Task 3, and MUStARD. Non-parametric retrieval resulted in a significant 9.87\% macro-F1 improvement on Twitter Indonesia Sarcastic compared to the original PMP method. Self-knowledge retrieval improves macro-F1 by 3.29\% on Semeval and by 4.08\% on MUStARD. These findings highlight the importance of context in enhancing LLMs performance in sarcasm detection task, particularly the involvement of culturally specific slang, references, or unknown terms to the LLMs. Future work will focus on optimizing the retrieval of relevant contextual information and examining how retrieval quality affects performance. The experiment code is available at: \url{https://github.com/wllchrst/sarcasm-detection_pmp_knowledge-base}. 
\end{abstract}


\section{Introduction}
\label{sec:introduction}
In the field of machine learning, natural language processing (NLP) tasks have been shown to be a crucial part of human life. NLP tasks revolve around processing text in a specific manner and receiving an output that can be useful, with examples such as text classification, text generation, information retrieval, and similar related tasks. In this study, we direct our focus toward text classification, specifically examining one of its key sub-tasks: sarcasm detection. Sarcasm detection, or as some call it verbal irony detection, is a task in NLP that automatically classifies text, and in extended forms includes images, audio, or video, as either sarcastic or not. Systems designed for sarcasm detection are becoming more important in the 21st century due to the growth of the use of sarcasm detection datasets caused by media usage of it, such as social media, television, and much more. Additionally, enhancing these automatic systems for sarcasm detection could become crucial in interpreting the real sentence meaning of a text. For example, a song review saying "The album is really incredible, it even made me sleep while listening to it" might be classified as something positive by a sentiment classification model of the word "incredible." However, upon further reading, the reviewer says that it made them sleep, which suggests that the album is boring.

Recent advances in machine learning studies and architectures have led to many exceptional performances across NLP tasks, such as sentiment and emotion classification. However, despite these improvements that have also impacted the overall performance of sarcasm detection, it still remains a challenging problem as it often relies on subtle contextual, pragmatic, or cultural cues. Large Language Models (LLMs) have shown tremendous capability in emulating human reasoning and completing specific tasks that have predefined rules \cite{llm-survey}. Multiple studies recently have shifted their focus toward methods for improving LLM performance, such as external information \cite{first-rag}, prompt techniques \cite{cot-prompting}, information retrieval technique \cite{ircot}, and related methods.

\begin{figure}
    \centering
    \includegraphics[width=\columnwidth]{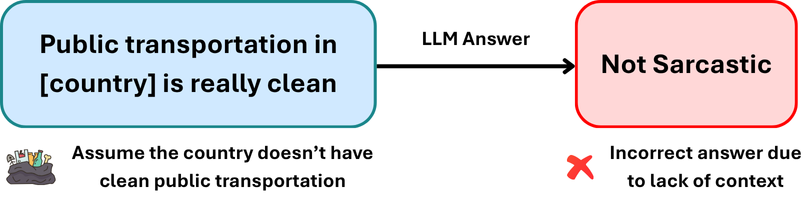}
    \caption{Extreme diversity and culture example}
    \label{fig:sd-without-context}
\end{figure}

The latest studies on sarcasm detection have benefited LLMs in their new method called Pragmatic Metacognitive Prompting (PMP) \cite{pmp}, which focuses on guiding LLMs to analyze and extract pragmatic information from a text. However, accurately interpreting a text requires a nuanced understanding of word meanings and their contextual relationships within the text. Unfortunately, due to the vast diversity of language and real-world context, this remains a major challenge for LLMs \cite{mallen}, as illustrated by the example in Figure \ref{fig:sd-without-context} for sarcasm detection. Hence, recent studies on LLMs have focused on incorporating non-parametric knowledge for tasks that are knowledge intensive, e.g., question answering and fact verification. Retrieval-Augmented Generation (RAG) has emerged to be the most popular topic in solving the problem, retrieving external information with the objective of improving accuracy and reducing hallucinations. Existing RAG studies are fairly scattered, with some focusing on the retrieval process \cite{gnn-rag} \cite{adaptive-rag}, while others examine the behavior of RAG \cite{rag-safe}.

Beyond the non-parametric knowledge, though accommodated with sufficient internal knowledge, models sometimes fail to flag which words or tokens need extra grounding. Such failures arise because internal knowledge is uneven and not reliably activated \cite{internal-uneven}. This could lead to a weaker analysis of the overall text for the final verdict of sarcasm detection. Recent studies have shown that eliciting the internal knowledge of LLMs can mitigate this issue by increasing its contextual awareness and reasoning quality. Such self-knowledge prompting not only helps a model determine when additional information may be needed, but also enhances its ability to reason more precisely even without relying on any external retrieval process \cite{internal-aware} \cite{internal-guide-retrieval}.

Building on this line of work, our study investigates the impact of non-parametric and internal knowledge awareness as context to primarily solve the limitation of using LLMs in a zero-shot manner when doing sarcasm detection. The non-parametric knowledge is retrieved based on semantically important words, which are identified using an LLM-based extraction. These words are then retrieved using Google Search API with the results cleaned by the LLM. Meanwhile, the internal knowledge is retrieved from the LLM itself with the important words being identified using Named Entity Recognition (NER) and Part-of-Speech (POS) tagging. All of these methods, from keyword extraction to context retrieval, are detailed in later parts of the paper. In addition, our study integrates the PMP method, which is then evaluated in English and Indonesian datasets, the latter consisting of traditional and culture-specific varieties of Indonesian. This combined method shows that when appropriate context is provided to LLMs, it plays an important role in helping them to gain deeper understanding of the text, therefore improving their ability to interpret and classify sarcastic text.

Contributions from this research are listed below:
\begin{itemize}
    \item Discovering the important role of non-parametric and internal knowledge awareness in sarcasm detection using LLMs, specifically through the state-of-the-art PMP method.
    \item Investigating the best course of action to identify words that are not fully understood by LLMs before performing information retrieval.
    \item Implementing simple information retrieval and producing a high margin of performance difference, revealing that a better information retrieval method can be highly beneficial.
\end{itemize}

The remainder of this paper is structured as follows. Section \ref{sec:related} reviews related works on text classification, sarcasm detection, NER, POS tagging, and RAG systems. Section \ref{sec:methods} explains in detail the methods used for key word extraction, information retrieval, and prompting techniques. Section \ref{sec:experiments} explains which datasets, models, and evaluation metrics are used for our experiment evaluation. Section \ref{sec:results} presents and analyzes each result of the experiment. Finally, Section \ref{sec:conclusion} concludes the paper and highlights its advantages, limitations, and directions for future work.

\section{Related Works}
\label{sec:related}

In this section, we will focus on the development of sarcasm detection, in which the explanation of earlier methods of text classification leading to sarcasm detection using Large Language Models (LLMs).

\subsection{Text Classification}
\label{sub:text-classification}
Text classification, the act of automatically classifying a text into a predefined set of labels, has been one of the most extensively researched areas of the earlier development of machine learning until now. Especially with today's growing influence of the internet and social media, it has been impacting the way machine learning is used for text classification tasks such as sentiment classification \cite{sentiment-classification}, emotion classification \cite{emotion-classification}, spam detection \cite{spam-detection}, sarcasm detection \cite{sarcasm-detection}, and much more. In addition to conventional tasks of text classification, recent studies have shifted focus toward emerging domains such as multilingual text classification \cite{multilingual-example}, multi-modal emotion classification \cite{multimodal-example}, and other similar classifications.

Early text classification tasks relied primarily on classical machine learning algorithms such as clustering \cite{clustering}, Naive Bayes \cite{naive-bayes}, and Support Vector Machines \cite{svm}, to name a few. These algorithms depended heavily on feature extraction, a process defined as converting text into numerical data, usually in the format of vectors, in order to be used as input for the algorithm. Some of the most frequently used methods for this process were TF-IDF \cite{tf-idf}, n-gram \cite{n-gram}, and bag-of-words \cite{bow}. While effective, these features often failed to take into account the semantic meaning and contextual relationships between words, focusing only on the word count of each sentence.

Nonetheless, these methods became an important building block for studies in text classification when researchers discovered neural networks. Neural networks enhance the feature extraction process by placing greater emphasis on the semantics and context of each word in a sentence. Some of their most used frameworks include Convolutional Neural Network (CNN) \cite{cnn}, Recurrent Neural Network (RNN) \cite{rnn}, and Recurrent Convolutional Neural Network (RCNN) \cite{rcnn}. These developments shifted the field toward deep models for text classification. As a result, one of the most significant innovations in neural networks of this decade was the Transformer framework \cite{transformer}, which laid the foundation for most modern language models, including BERT \cite{bert}  and LLMs, such as GPT \cite{gpt} and Gemini \cite{gemini}, to name a few.

The rapid advancement of text classification has effectively addressed what can be described as "System I" tasks, those that are fast, unconscious, and intuitive. These tasks have been argued to be successfully solved, including sentiment analysis, emotion classification, and spam detection. In contrast, "System II" tasks require slower, deliberate, and multi-step cognitive processes, such as sarcasm detection, which remains an unsolved challenge \cite{sarcasm-bench}.

\subsection{Sarcasm Detection}
Sarcasm detection remains one of the most challenging tasks in text classification, despite often being formulated as a simple binary problem (sarcastic or not sarcastic). Recent sarcasm detection studies have also expanded toward multi-label \cite{i-sarcasm} and multi-modal \cite{mustard} settings. These approaches aim to capture the deeper pragmatic and contextual cues of sarcasm, which often rely not only on text but also on tone of voice, facial expressions, and situational context.

Recent studies of sarcasm detection have focused on dataset creation and augmentation. Some works even employ deep learning frameworks, such as the Reverse Generative Adversarial Network, to handle the dataset augmentation \cite{suhartono2023feature}. Other studies emphasize dataset construction, for example, collecting sarcastic lines from TV shows \cite{mustard} or by using web scraping methods \cite{suhartono2024idsarcasm}.

\subsubsection{Pre-trained Language Models}

Consistent with the development of neural networks, recent studies on sarcasm detection have also adopted models based on transformer, such as BERT \cite{sarc-bert} and RoBERTa \cite{sarc-roberta}. Building on the strong contextual representation capabilities of these models, researchers have started to explore more advanced Pre-trained Language Models (PLMs) that can be fine-tuned or adapted specifically for sarcasm detection using text. For instance, Dual-Channel Network (DC-Net) \cite{dual-channel-plm} was designed solely for sarcasm detection by modeling distinct sentiment cues from text segments before combining them for classification. Similarly, SarcPrompt \cite{prompt-tuning-plm} introduces prompt-based fine-tuning to guide PLMs in identifying sarcasm polarity more accurately.

\subsubsection{Large Language Models}
Although PLMs have proven effective for sarcasm detection, recent advances in Large Language Models (LLMs) have also shown remarkable progress in this area, along with an increasing number of studies investigating their effectiveness and applications. Their ability to capture nuanced contextual and pragmatic cues has allowed researchers to approach sarcasm detection from a more reasoning-oriented perspective. Initially focused on text generation, LLMs now excel in multiple tasks such as reasoning \cite{llm-reasoning}, code generation \cite{llm-code}, and question answering \cite{llm-qa}.

\textbf{Prompting Methods} \hspace{2mm} A key feature that enables LLMs to perform diverse tasks is prompting, the process of guiding the model's reasoning or behavior through carefully constructed instructions. In the context of text classification, prompting serves not only to map input text to predefined labels but also to influence how the model interprets context, structure, and intent. Common prompting strategies for text classification include the following:
\begin{itemize}
    \item Zero-shot prompting: provides LLMs with a natural-language prompt describing the classification goal to directly predict the appropriate label \cite{zero-shot}.
    \item Few-shot prompting: introduces LLMs with labeled examples within the prompt to demonstrate the desired mapping before classifying new inputs \cite{few-shot}.
    \item Chain-of-Thought (CoT): uses a series of intermediate reasoning steps to enhance the reasoning capabilities of LLMs in tasks such as common sense inference and arithmetic problems \cite{cot-prompting}.
\end{itemize}

\textbf{Pragmatic Metacognitive Prompting} \hspace{2mm} Beyond standard prompting methods such as zero-shot, few-shot, and chain of thought prompting, recent studies in sarcasm detection have focused on finding more effective prompting strategies. So far, Pragmatic Metacognitive Prompting (PMP) \cite{pmp} has shown promising progress with higher performance in every metric compared to PLMs. PMP itself is built on top of Metacognitive Prompting (MP) \cite{mp}, a framework that introduces the concept of preliminary analysis followed by a reflection stage before giving the final judgment. PMP adopts this same infrastructure, but extends it by embedding simplified pragmatic components into both stages. The first stage focuses on extracting and interpreting pragmatic elements from the text, while the second stage reflects on the earlier analysis to form a deeper reasoning about the meaning and intent behind the statement. This process allows the model not only to evaluate what is said, but also to understand what is implied, which improves its ability to detect sarcasm and indirect meaning in the text.

\begin{table}[h]
\caption{Sarcasm detection performance from previous studies}
\centering
\renewcommand{\arraystretch}{1.3}
\begin{tabular}{l l c c}
\hline
\textbf{Method} & \textbf{Model} & \textbf{Accuracy} & \textbf{Macro-F1} \\
\hline
\multicolumn{4}{l}{\textit{MUStARD Dataset}} \\
\hline
PMP & GPT-4o & 0.7942 & 0.7765 \\
PMP & LLaMA-3-8B & 0.5358 & 0.5469 \\
PLMs & RoBERTa & 0.8680 & 0.8770 \\
PLMs & DistilBERT & \textbf{0.8700} & \textbf{0.8770} \\
\hline
\multicolumn{4}{l}{\textit{SemEval 2018 Dataset}} \\
\hline
PMP & GPT-4o & \textbf{0.8668} & \textbf{0.8318} \\
PMP & LLaMA-3-8B & 0.7821 & 0.7765 \\
PLMs & RoBERTa & 0.7500 & 0.7200 \\
PLMs & DC-Net-RoBERTa & 0.7090 & 0.6870 \\
\hline
\multicolumn{4}{l}{\textit{Twitter Indonesian Dataset}} \\
\hline
Zero-shot & mT0XL & 0.2494 & 0.3989 \\
PLMs & XLM-RLarge & \textbf{0.8885} & \textbf{0.7692} \\
\hline
\end{tabular}
\label{tab:previous_result}
\\[2pt]
\footnotesize{\textit{Note:} Macro-F1 denotes the \textbf{macro-averaged F1-score} across all classes.}
\end{table}

However, PMP does not have leverage on all datasets, showing state-of-the-art performance in SemEval 2018 \cite{semeval} but inferior results when evaluated in MUStARD \cite{mustard}. This variation stems from a broader limitation of LLMs, namely their reliance on parametric knowledge. Because PMP operates in a zero-shot setting without access to external context, its effectiveness decreases when sarcasm deviates from general linguistic or cultural norms. The comparison of PMP and PLMs performance can be seen in Table \ref{tab:previous_result}.

\subsection{Named Entity Recognition}
Named Entity Recognition (NER) is a fundamental task in Natural Language Processing (NLP) that aims to identify entities of interest in text, such as people, locations, and organizations. This task plays an important role in guiding the semantic understanding of text. \cite{ner-definition}. Similar to other NLP tasks, early approaches to NER relied heavily on handcrafted features and statistical models. With the advancement of deep neural networks across various machine learning domains, as discussed in Subsection \ref{sub:text-classification}, these architectures have also become the foundation for modern NER systems. A recent state-of-the-art framework, BiLSTM-CRF \cite{ner-bilstm}, achieved strong performance and outperformed traditional NER methods. However, these architectures often under-perform in transferability compared to pre-trained transformer-based models such as BERT \cite{ner-bert}, which is particularly relevant to this study due to its ability to generalize across domains and languages. At the same time, despite the success of transformer-based architectures, smaller neural network models remain valuable due to their computational efficiency. A popular NLP library in Python, spaCy, adopts a lightweight CNN-based architecture for its NER and POS tagging components, offering an efficient alternative that integrates well into larger NLP systems \cite{spacy}.

\subsection{POS Tagging}
Part-of-Speech (POS) tagging assigns a syntactic category to each word in a text and is a fundamental task in NLP. Similar to NER, POS tagging helps models understand the structure and function of words in a sentence. Early approaches relied on classical machine learning algorithms such as Naive Bayes, Hidden Markov Models (HMM), and Conditional Random Field (CRF) \cite{pos-definition}. With the rise of deep learning, sequence models, including Gated Recurrent Unit (GRU), Recurrent Neural Network (RNN), and Bi-directional Long Short Term Memory (LSTM) \cite{pos-dl} \cite{pos-dl-hindi}, have become dominant in the field. More recently, studies have shifted focused on identifying the effectiveness of POS tagging using LLMs \cite{pos-llm} and transformer-based models like BERT \cite{pos-bert}.

\subsection{Retrieval-Augmented Generation}
One major issue with LLMs is hallucination \cite{llm-hallu}. Previous works have shown that the parametric knowledge of LLMs can be unreliable in open-domain question answering, and that performance varies depending on the popularity of the entity being queried; in particular, less common entities tend to increase the likelihood of hallucination \cite{mallen}. Retrieval-Augmented Generation (RAG) addresses this by augmenting model prompts with non-parametric knowledge retrieved from external sources, which improve factuality and reduce hallucination. In practice, retrieval can be implemented with sparse methods (e.g., BM25), dense retrievers like Dense Passage Retrieval (DPR), or hybrid pipelines that combine both. A recent work known as Interleaving Retrieval with Chain of Thought prompting (IRCoT) \cite{ircot} has also explored non-parametric knowledge retrieval but combines it with CoT prompting technique, which has shown a tremendous result in knowledge augmentation \cite{cot-knowledge}.

\subsection{Context-Awareness and Internal Knowledge}
Researchers have found that LLMs do not always recognize which parts of an input need extra background or deeper reasoning. This can result in weak performance in tasks that require subtle interpretation \cite{llm-fact-boundary}. One line of work has extracted signals from the model itself, such as uncertainty or “do I know this?” responses, to decide when to look up external information. For example, a study proposed a method in which the model checks its internal knowledge and only prompts retrieval when needed \cite{internal-guide-retrieval}. Another study asked LLMs to generate short internal summaries about entities or concepts present in the input, and showed that this helped even without retrieving large external corpora \cite{internal-aware}. Additional work has used the internal hidden states of transformer layers to estimate the model’s self-awareness of its factual bounds, and linked this measurement to performance and retrieval decisions \cite{llm-fact-measure}. Overall, this literature highlights that improving a model’s awareness of its own knowledge state is a useful complement to external retrieval and can itself enhance reasoning quality.

\section{Methodology}
\label{sec:methods}

To solve the problem of specific linguistic norms in texts for sarcasm detection, we propose a method that is inspired by Retrieval-Augmented Generation (RAG), in which our method focuses on identifying, retrieving, and providing information that the model may not understand. In this section, we explain the retrieval step and the prompting method used in this study.

\subsection{Retrieval Method}
Our information retrieval process begins with the identification of relevant words, which are later used as keywords during the retrieval process.

\subsubsection{Keyword Extraction}
Keyword quality highly matters for the performance of the LLM. Too many keywords increase query size, add noise, and may push the input past token or cost limits. Therefore, the extraction of keywords from the text should only include words that most improve the model's understanding. Our method proposes two extraction approaches, namely Token Tagging and LLM-Based.

\begin{figure}
\centering
\includegraphics[width=\columnwidth]{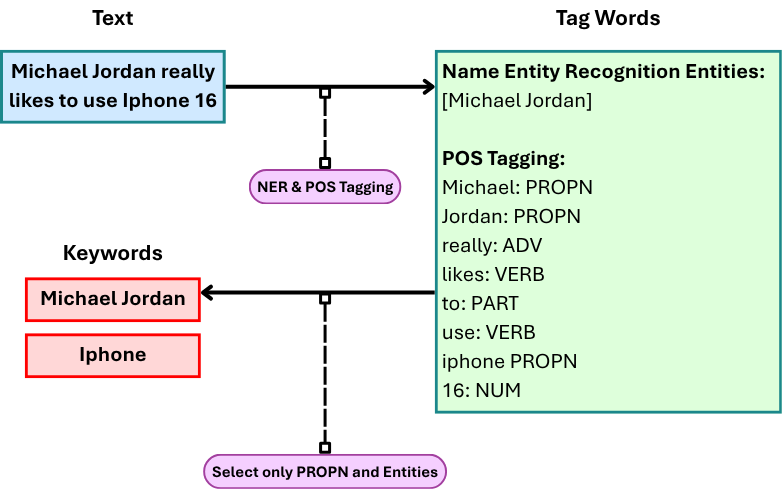}
\caption{Token Tagging Extraction}
\label{fig:token-tagging-extraction}
\end{figure}

\textbf{Token Tagging} \hspace{2mm} Let us first define our approach in Token Tagging, in which our approach uses Named Entity Recognition (NER) and Part-of-Speech (POS) tagging as illustrated in Figure \ref{fig:token-tagging-extraction}. Given a task of sarcasm detection of the text T, T is first processed by using a pipeline that tags each word, or token, with their respective entity type and part of speech. We focus only on tokens identified as entities or proper nouns. Entities refer to real-world objects with distinct types such as people, locations, and organizations, for example, "Elon Musk", "Paris", or "Google". Proper nouns are included as well due to NER models occasionally failing to recognize certain entities; proper nouns often capture those missed cases. Our extraction is restricted to these tokens because they carry strong contextual meaning, especially in domain-specific cases, helping the model interpret the text more accurately.

\begin{figure}
\centering
\includegraphics[width=\columnwidth]{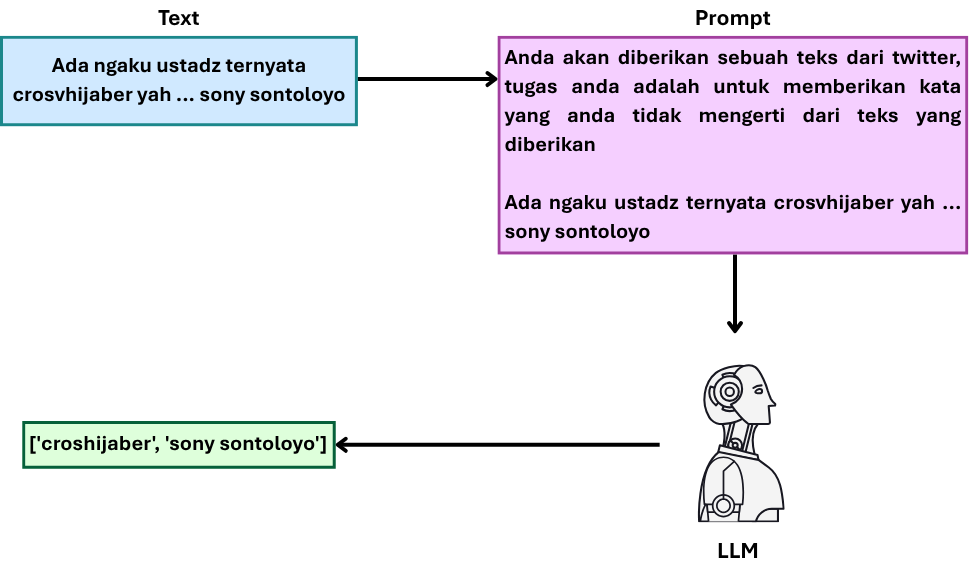}
\caption{LLM Based Extraction}
\label{fig:llm-based-extraction}
\end{figure}

\textbf{LLM-Based} \hspace{2mm} The reasoning behind using an LLM to extract keywords is inspired by the IRCoT method \cite{ircot}, which employs LLMs to guide the retrieval process. However, IRCoT focuses primarily on question-answering tasks, specifically multi-hop retrieval, where the LLM is used to iteratively refine queries to support reasoning. In contrast, our method leverages the LLM in a different capacity. Rather than generating refined queries, we utilize the LLM to automatically extract words from raw text. This method was initially used for the Indonesian Twitter dataset, which contains a lot of slang words in various languages, including Indonesian and Javanese, which makes this particular approach preferable to Token Tagging. In our approach, the input text T is provided to the LLM along with an additional prompt that instructs the model to identify and return a list of words it does not recognize or cannot infer from prior knowledge. These unknown or ambiguous terms are hypothesized to carry domain specificity or contextual importance, which makes them strong candidates for keyword information retrieval. The details of the prompt used to guide the LLM in this extraction process are illustrated in Figure \ref{fig:llm-based-extraction}.

\subsubsection{Word-Information Retrieval}
After selecting keywords that require additional context, we send them to a word-information retrieval pipeline. Our approach implements two complementary retrieval strategies, first using Google Search API and LLM-based cleaning, and second using an LLM-only approach.

\begin{figure*}
\centering
\includegraphics[width=\textwidth]{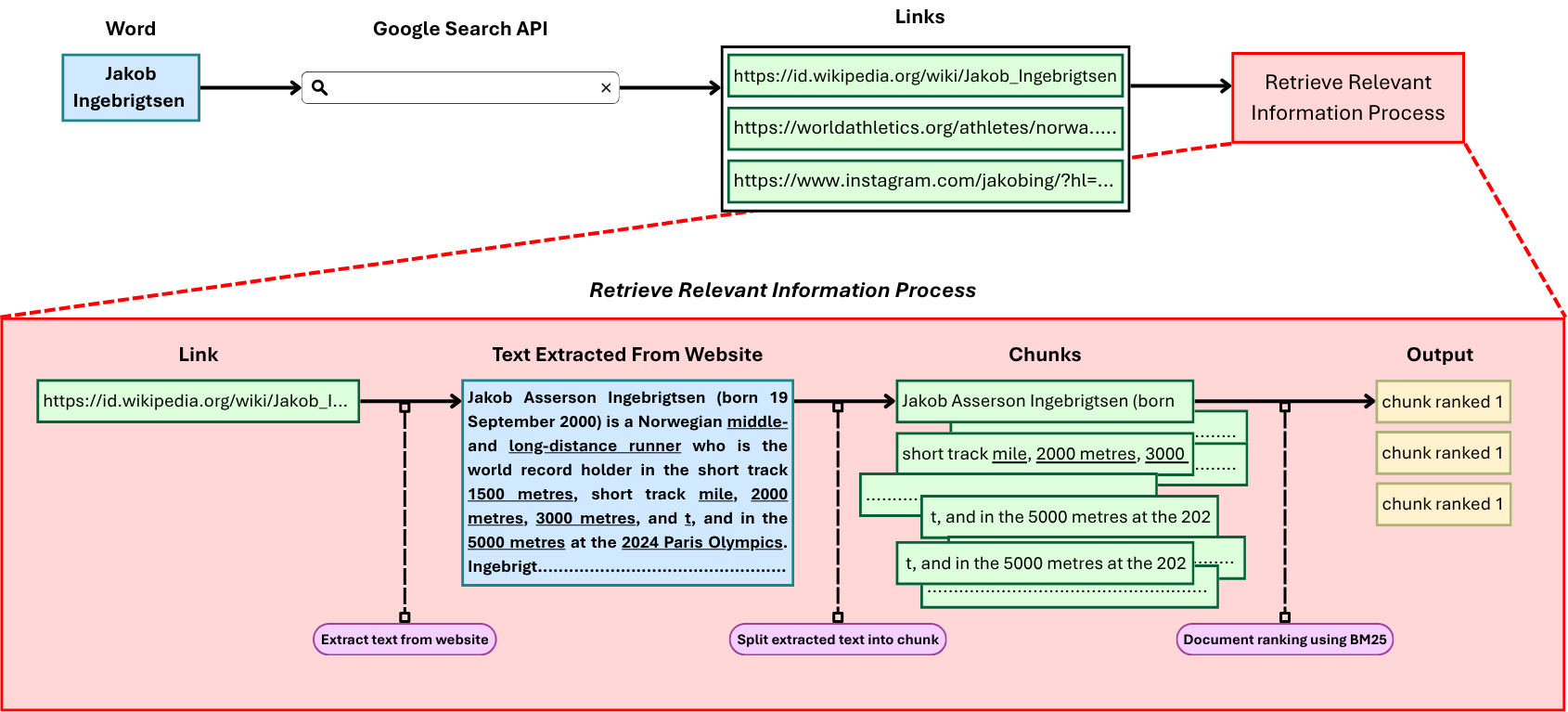}
\caption{Retrieve Link Process}
\label{fig:retrieve-information-internet}
\end{figure*}

\textbf{Google Search API + LLM} \hspace{2mm} Each extracted keyword is used as a query to retrieve candidate links and snippets from the web. We utilize the Google Search API for this step. This allows the system to gather real-world usage examples and contextual knowledge that may not be present in the knowledge of the LLM.

However, raw search snippets often contain noise, redundancy, or irrelevant information. Therefore, after retrieving the results, we split the retrieved documents into manageable text chunks and score them with BM25 to select the most lexically relevant passages. BM25 requires low computational cost and is effective for definitional matches, which are useful when the meaning of a keyword is expressed explicitly on web pages. The link retrieval and document ranking steps are illustrated in Figure \ref{fig:retrieve-information-internet}. The BM25 ranking function is defined as:

\begin{equation}
\label{eq:bm25}
\resizebox{0.90\hsize}{!}{$
\text{BM25}(q, D) =
\sum_{i=1}^{n} IDF(q_i) \cdot 
\frac{f(q_i, D) \cdot (k_1 + 1)}{f(q_i, D) + k_1 \cdot \left(1 - b + b \cdot \frac{|D|}{\text{avgDL}}\right)}
$}
\end{equation}

where the inverse document frequency $IDF(q_i)$ is defined as:

\begin{equation}
IDF(q_i) = \ln \left( \frac{N - n(q_i) + 0.5}{n(q_i) + 0.5} + 1 \right)
\label{eq:bm25idf}
\end{equation}

\noindent The variables in Equation \ref{eq:bm25} and \ref{eq:bm25idf} are:
\begin{itemize}
  \item $q$ is the query containing terms $q_1, q_2, ..., q_n$.
  \item $D$ is a document.
  \item $f(q_i, D)$ is the term frequency of term $q_i$ in document $D$.
  \item $|D|$ is the document length in words.
  \item $\text{avgDL}$ is the average document length across the collection.
  \item $N$ is the total number of documents in the collection.
  \item $n(q_i)$ is the number of documents containing term $q_i$.
  \item $k_1$ and $b$ are hyperparameters, where typically $k_1 = 1.2$ and $b = 0.75$.
\end{itemize}

After ranking, we pass the highest ranked chunks to an LLM for refinement. The LLM is prompted to summarize the chunks, filter out irrelevant content, and extract key semantic meanings. For every chunk, the final output is a short definition of itself. This hybrid method is especially useful when the LLM itself lacks domain-specific parametric knowledge or other terms not covered well by the model.

\textbf{LLM-only} \hspace{2mm} In the LLM-only approach, the extracted keywords are provided as input to the LLM to generate a contextual definition for each keyword. The definition is based solely on its internal parametric knowledge, not relying on external knowledge sources. The process begins by providing the keyword as input to the LLM, which then generates a contextual definition based solely on its internal parametric knowledge. Although the accuracy of the generated definitions can vary depending on the keyword and the prior training of the model, this approach offers significant advantages in terms of efficiency, as it reduces both processing time and computational cost.

\subsubsection{Extraction-Retrieval Pairing}
\label{subsub:extraction-retrieval-pairing}
Between those two methods of keyword extraction and word-information retrieval methods, we provide a short justification for which pairings we evaluate and why those choices best match the dataset properties and practical constraints.

\textbf{Token Tagging + LLM-only} \hspace{2mm} For the first pairing, we use Token Tagging for keyword extraction and LLM-only word-information retrieval. This pairing is our method for adding internal knowledge context, specified for datasets that do not rely on heavy culture-specific slang or local references, for example, MUStARD, which contains mainly standard English captions. We chose this pairing because when keywords come from a separate, structured extractor, asking the model to define them uses its internal knowledge without creating a circular process. If the model both selects the tokens and then tries to define them, in this case pairing up LLM keyword extraction and retrieval, the workflow can simply reinforce its own mistakes. Some prior work has used the self-asking pattern for internal knowledge to boost context awareness, but we expect cleaner and more reliable context when the model is grounded by an external extractor like Token Tagging.

\textbf{LLM-based + Google Search API} \hspace{2mm} The second pairing uses LLM-based keyword extraction and Google Search API word-information retrieval. This pairing is our method for adding non-parametric knowledge context, particularly designed for the Indonesian Twitter dataset, which contains many domain-specific slang and terms that spaCy models cannot reliably detect as named entities or proper nouns, so Token Tagging would miss many candidate tokens. Instead, we ask the LLM directly which tokens it does not recognize. Those flagged tokens are then used to query the Google Search API for non-parametric knowledge, avoiding the LLM-only retrieval approach, which clearly would not have any information about the tokens. Note that we do not apply Google Search API retrieval to datasets made up mainly of standard English captions, for example, MUStARD. In those cases, the LLM already contains the relevant background knowledge, and adding web retrieval tends to introduce irrelevant snippets and noise rather than improve understanding.

\subsection{Prompting Method}
In our prompting methods, we involve three primitive prompting components, such as Pragmatic Metacognitive Prompting (PMP), few-shot, and word-information prompting. From there, two hybrid pipeline templates are produced from combinations between them, with a final total number of five prompting variants, such as a PMP baseline and four hybrids.

\subsubsection{Primitives Prompting Components}
This part describes the three primitive prompting components we use as building blocks, namely PMP, few-shot, and word-information prompting.

\textbf{Pragmatic Metacognitive Prompting} \hspace{2mm} Our study uses the PMP pipeline for the baseline and skeleton, with two LLM calls per input $X$. The first call, denoted $P_{1}$, asks the model to understand $X$ generally and produce an initial pragmatic analysis, denoted $A_{1}$. The second call, denoted $P_{2}$, reflects on the analysis of the first call to produce a more structured pragmatic analysis, containing factors from a pragmatic viewpoint such as the implicature, presupposition, intent, polarity, pretense, and potential meanings individually, denoted $A_{2}$. From that analysis, the second call also verdicts at the end whether the context is sarcastic or not with a final label, $y$, of "YES" or "NO".

\textbf{Few-Shot Prompting} \hspace{2mm} In this approach, a small set of illustrative examples is provided to guide the reasoning process of the LLM. Rather than serving solely as reference points for producing correct labels or mappings, these examples function as cognitive cues that help the model enhance its internal thought process. These short examples, denoted $F$, are made manually and tailored to the dataset theme, with each typically containing an input, the structured analysis, and the final label. The number of examples, parameter k, is intentionally small to maintain clarity and reduce noise. We used k = 2 in our experiments.

\textbf{Word Information Prompting} \hspace{2mm} The word-information retrieval returns contextual definitions, using either Google Search API with LLM-based keyword extraction, denoted $W_{g}$, or LLM-only approach with Token Tagging keyword extraction, denoted $W_{l}$. These retrieved definitions are included in the prompt alongside the input to help the model better understand for further analysis. When the model sees concise clarifications for slang, named-entities, or culture-specific terms it might not otherwise know, further analysis becomes less noisy and more accurate. Each definition is kept to one or two sentences so the prompt stays compact and readable.

\subsubsection{Notations and Pipeline Overview}
Below we define formal notations used in the prompting pipelines and present the PMP skeleton as building blocks for the two hybrid pipeline templates.

\noindent\textbf{Notations:}
\begin{itemize}
  \item[$X$] input text.
  \item[$P_{1}$] first call of PMP.
  \item[$P_{2}$] second call of PMP.
  \item[$A_{1}$] preliminary pragmatic analysis\\(output of $P_{1}$).
  \item[$A_{2}$] structured reflection\\(output of $P_{2}$).
  \item[$K_{t}$] keywords extracted from $X$ using Token Tagging.
  \item[$K_{l}$] keywords extracted from $X$ using LLM-based.
  \item[$W_{l}$] word-info produced by LLM-only approach\\(RetrieveLLM($K_t$).
  \item[$W_{g}$] word-info produced by Google Search API $\to$ LLM\\(RetrieveGoogle($K_l$)).
  \item[$W$] chosen word-info block\\$\in\{\varnothing, W_{l}, W_{g}\}$.
  \item[$F$] few-shot demonstrations\\(k examples).
  \item[$y$] final label\\$\in\{\text{YES},\text{NO}\}$.
  \item[$\oplus$] concatenation into prompt\\(append to user or system prompt).
\end{itemize}

\noindent\textbf{PMP skeleton:}
\begin{itemize}
  \item[] $A_{1} = P_{1}(X)$
  \item[] $(A_{2}, y) = P_{2}(A_{1})$
\end{itemize}

\noindent\textbf{First hybrid template (PMP + Word-Info):}
\begin{itemize}
  \item[] $A_{1} = P_{1}(X \oplus W)$
  \item[] $(A_{2}, y) = P_{2}(A_{1})$
\end{itemize}

\noindent\textbf{Second hybrid template (PMP + Word-Info + Few-shot):}
\begin{itemize}
  \item[] $A_{1} = P_{1}(X \oplus W)$
  \item[] $(A_{2}, y) = P_{2}(A_{1} \oplus F)$
\end{itemize}

\subsubsection{Hybrid Prompting Pipelines}
This part presents the two hybrid pipeline templates, each realized with Google Search API or LLM word-information. Combining these templates with the PMP baseline yields five prompting variants evaluated.

\begin{figure}
\centering
\includegraphics[width=\columnwidth]{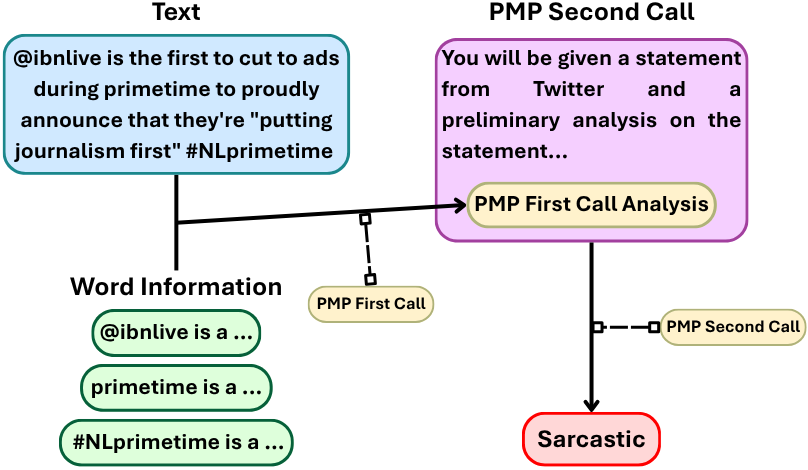}
\caption{First Hybrid Prompting Template}
\label{fig:pmp_wordi}
\end{figure}

\textbf{First Hybrid Template (PMP + Word-Info)} \hspace{2mm} In this first hybrid variant, the pipeline integrates the word-information block ($W$) into the PMP process, which can be seen in Figure \ref{fig:pmp_wordi}. With PMP having two calls, putting the block in the first call ($P_{1}$) would be more suitable, as it is the model's first time encountering and understanding the input. The goal here is to give the model more grounding on specific terms that may not exist in its prior knowledge, such as slang, uncommon entities, or local cultural phrases. Introducing the block in the second call ($P_{2}$) would not just already be too late, but could also confuse the reflection process and add unnecessary noise. By providing the word information early, the model can process every keyword and meaning together with the input from the start, allowing the $P_{2}$, the reflection stage, to focus purely on deeper pragmatic reasoning instead of fixing misunderstandings about the words themselves.

Regarding whether it should be placed in the user or system prompt, we put it in the user prompt since it is also part of the input that the model needs to process. System prompts are filled with instructions, shaping the model's behavioral and thinking guide, while user prompts are read by the model as "what to process". Therefore, $W$ fit better in the user prompt to be processed together with the main text, helping the model understand the meaning contextually before continuing to analyze it pragmatically.

\begin{figure}
\centering
\includegraphics[width=\columnwidth]{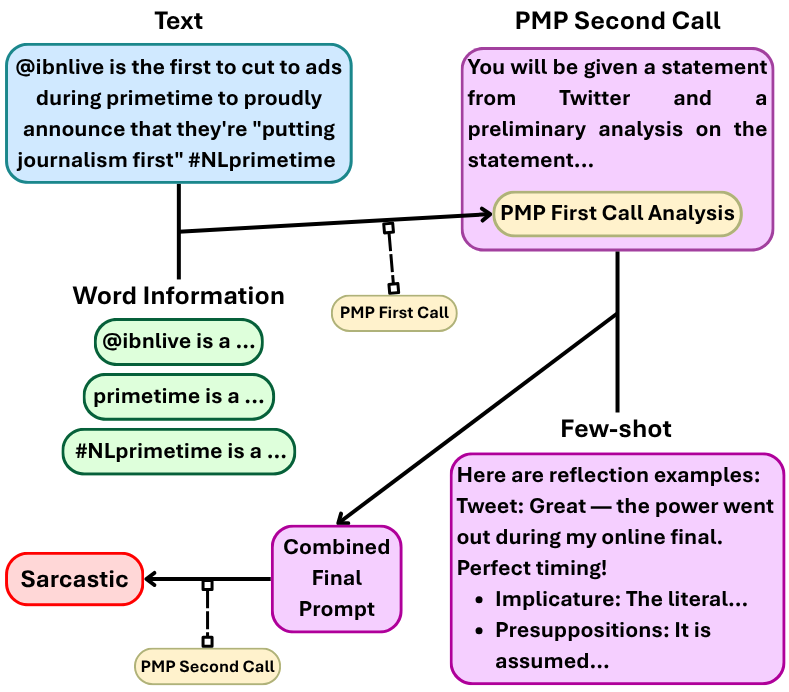}
\caption{Second Hybrid Prompting Method}
\label{fig:pmp_wordi_fewshot}
\end{figure}

\textbf{Second Hybrid Template (PMP + Word-Info + Few-Shot} \hspace{2mm} For another hybrid variant, we add few-shot examples ($F$) to the first hybrid variant as illustrated in Figure \ref{fig:pmp_wordi_fewshot}. $F$ are placed on the system prompt, not on the user prompt. As explained before, the system prompt mainly controls how the model behaves and thinks. They bias internal heuristics and the model's way of combining factors, not merely the output tokens, a big advantage in improving the PMP analysis.

Then, with PMP having two calls, it is unnecessary to put $F$ integration at the first call ($A_{1}$) due to it's task to only do a quick and general pragmatic scan. Throwing $F$ here only adds noise that could impact the final verdict later on. However, the second call ($A_{2}$) is a structured step, a complex analysis to closely inspect each pragmatic factor and then combine those pieces into a final verdict. This step demands careful, factor-by-factor reasoning, not because it is more complex in an abstract sense, but because it requires precise mapping from the analysis to a label. $F$ could prove to be useful here.

\begin{table}
\centering
\renewcommand{\arraystretch}{1.3}
\caption{Pipeline Experiment Settings}
\label{tab:pipelines}
\begin{tabular}{|p{2cm}|p{1.5cm}|p{1.5cm}|p{1.5cm}|}
\hline
\textbf{Setting} & \textbf{Word-Info (LLM-only $W_{l}$)}& \textbf{Word-Info (Google Search API $W_{g}$)}& \textbf{Few-shot ($F$)}\\ \hline
\rowcolor[HTML]{CBCEFB} 
PMP & & & \\ \hline
\rowcolor[HTML]{FFFFC7} 
PMPWL& V & & \\ \hline
\rowcolor[HTML]{FFCCC9} 
PMPWG& & V & \\ \hline
\rowcolor[HTML]{FFFFC7} 
PMPWL-FS& V & & V \\ \hline
\rowcolor[HTML]{FFCCC9} 
PMPWG-FS& & V & V \\ \hline
\end{tabular}
\end{table}

\textbf{Final Five Prompting Variants} \hspace{2mm}
The two hybrid templates each produce two variants, one from Google Search API word-information retrieval with LLM-based keyword extraction, and one from LLM-only word-information retrieval with Token Tagging keyword extraction. From this, we have four hybrid variants, with a total number of five prompting variants when combined with the PMP baseline as categorized in Table \ref{tab:pipelines}. Here are their formulas:
\begin{itemize}
    \item PMP: $y = P_{2}(P_{1}(X))$
    \item PMPWL: $y = P_{2}(P_{1}(X \oplus W_{l}))$
    \item PMPWG: $y = P_{2}(P_{1}(X \oplus W_{g}))$
    \item PMPWL-FS: $y = P_{2}(P_{1}(X \oplus W_{l})) \oplus F)$
    \item PMPWG-FS: $y = P_{2}(P_{1}(X \oplus W_{g})) \oplus F)$
\end{itemize}

\section{Experiments}
\label{sec:experiments}

In this section, we explain the datasets, models, evaluation metrics, and implementation details that we used. All experiments and evaluations executed in this study, including keyword extraction, word-information retrieval, and pipeline evaluations, were conducted on a machine equipped with an Nvidia RTX 3090 GPU and 24 GB of VRAM. Due to hardware constraints, we were only able to run experiments on small-scale models. This prevented the evaluation of larger models and consequently limited our ability to investigate the impact of context in higher-capacity architectures.

\subsection{Datasets}
\begin{figure*}
\centering
\includegraphics[width=\textwidth]{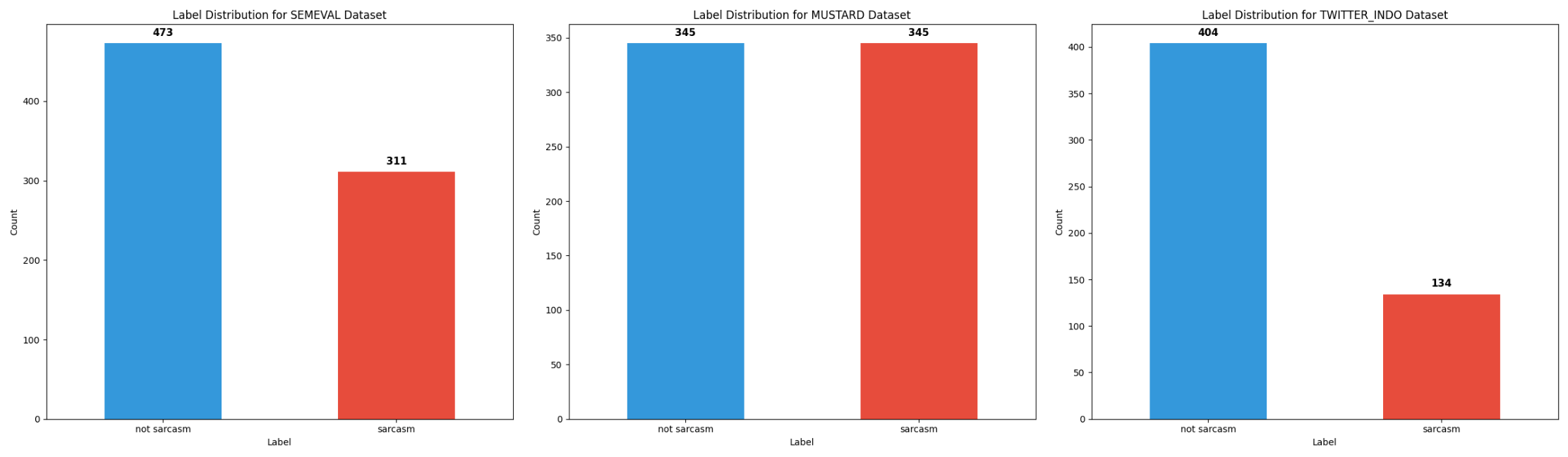}
\caption{Dataset Distribution (Test)}
\label{fig:dataset-ditribution}
\end{figure*}

To evaluate the impact of the word-information context accurately, we conduct evaluations of our pipelines using three different datasets, with three different themes. The datasets are \textbf{1) SemEval-2018 Task 3}, an English twitter dataset where sarcastic or ironic tweets are collected by the hashtags, \textbf{2) MUStARD}, a multi-modal dataset created from three different TV shows: Friends, The Golden Girls, and Sarcasmaholics Anonymous, \textbf{3) Indonesian Twitter}, a labeled twitter dataset curated from March 2013 to February 2020, which consists of multiple languages including Indonesian, Javanese, and other traditional languages. The distribution of the datasets is illustrated in Figure \ref{fig:dataset-ditribution}. Unfortunately, the class distribution of the SemEval and Indonesian Twitter datasets is imbalanced, resulting in some evaluation metrics being less reliable. The reason for choosing these three datasets is to compare our results with the original PMP study, which uses SemEval-2018 Task 3 and MUStARD. In addition, we chose the Indonesian Twitter dataset to further explore the impact of non-parametric knowledge context when detecting sarcasm in text, which is a more extreme linguistic challenge for LLMs.
The datasets are processed in the same way as in the PMP study, where Semeval-2018 and the Indonesian Twitter dataset, although the latter was not included in the study, are already in the desired form, having the text (tweet) and the sarcastic label (binary) for each line. Meanwhile, for MUStARD, while being a multi-modal dataset, it also offers a JSON format version which includes the text (dialogue), the sarcastic label (binary), and previous dialogues. We follow the same pre-processing method used in the PMP study, which was to combine all previous and target dialogue into one line, with the target being in curly brackets. Also, we used the test split version for the Indonesian Twitter dataset

\subsection{Models} NER and POS tagging are done using a spaCy model, specifically \textbf{en\_core\_web\_sm} due to the efficiency of the model in utilizing CPU computation capabilities. We decided to use this small model because the performance is already high based on the documentation on their website. Then for the LLMs, as mentioned previously, we are limited to small models, therefore open-source models \textbf{Qwen-3}, \textbf{Llama-3}, and additionally \textbf{Llama-3.1}, each with 8 billion parameters, are used for keyword extraction, information retrieval, and pipeline evaluation. These two models, among other 8-billion-parameter models, exhibit top-tier reasoning and knowledge performance, suitable for sarcasm detection. Qwen-3 is also trained across 119 languages and achieves top-rank multilingual benchmark results, an advantage for our Indonesian Twitter dataset. In addition, previous studies, including the PMP work, also utilized Llama-3, which we can also use for comparison with additionally Llama-3.1 due to its better reasoning.
For the spaCy model, we ran it locally on our machine. The LLM models are run through Ollama and use the model endpoints directly. Across all experiments, we did not change any settings. No temperature, top\_p, or decoding options were set and were left at default. Every model task call, such as keyword extraction, context retrieval, and pipeline execution, used the model defaults provided. This keeps behavior consistent across datasets, models, and pipeline variants but means results reflect the default decoding regime of the models rather than a tuned configuration, which is done to focus on the performance impact of additional context and methods.

\subsection{Evaluation Metrics} To evaluate the performance of our pipeline, we use standard classification metrics such as accuracy, precision, recall, and F1-score. However, because of the imbalanced dataset class distribution, the calculation uses the macro average which calculates the metric independently for each class, then takes the unweighted mean across all classes.

\subsection{Implementation Details}
The evaluation is conducted under three experimental settings, with a total of five different pipelines as summarized in Table \ref{tab:pipelines}. Word-info (LLM-only $W_{l}$) refers to the use of Token Tagging for keyword extraction, followed by context retrieval using an LLM. This setting is used especially for English-themed datasets, such as Semeval-2018 and MUStARD. Word-info (Google Search API $W_{g}$) denotes keyword extraction using LLM-based approach, followed by context retrieval using the Google Search API, with the retrieved information subsequently summarized by an LLM. This setting is used specifically for culture-specific datasets, such as the Indonesian Twitter dataset and Semeval-2018. The pairings of keyword extraction and context retrieval have been reasoned in Subsubsection \ref{subsub:extraction-retrieval-pairing}.

\begin{table*}[h]
\centering
\caption{Evaluation results across different datasets, models, and pipelines.}
\label{tab:evaluation_all}
\begin{adjustbox}{max width=\textwidth}
% \sisetup{round-mode=places, round-precision=4, table-format=1.4}
\begin{tabular}{
    @{}
    l l 
    S[table-format=1.4] 
    S[table-format=1.4] 
    S[table-format=1.4] 
    S[table-format=1.4] 
    % S[table-format=1.4, table-model-setup=\bfseries] 
   % S[table-format=1.4] 
    % S[table-format=1.4, table-model-setup=\bfseries]
    @{}
}
\toprule
\textbf{Model} & \textbf{Method} & {\textbf{Accuracy}} & {\textbf{Precision}} & {\textbf{Recall}} & {\textbf{Macro-F1}} \\
\midrule
\addlinespace
\multicolumn{6}{l}{\itshape Indonesian Twitter Dataset} \\
\midrule
\multirow{3}{*}{Qwen-3-8B}
  & PMP         & 0.6078 & 0.5189 & 0.5219 & 0.5173 \\
  & PMPWG       & 0.6673 & 0.6094 & \textbf{0.6363} & 0.6108 \\
  & PMPWG-FS    & 0.6989 & \textbf{0.6121} & 0.6225 & \textbf{0.6160} \\
\addlinespace
\multirow{3}{*}{Llama-3-8B}
  & PMP         & \textbf{0.4963} & 0.5324 & 0.5424 & 0.4794 \\
  & PMPWG       & 0.4907 & \textbf{0.5523} & \textbf{0.5661} & \textbf{0.4814} \\
  & PMPWG-FS    & 0.4665 & 0.5342 & 0.5426 & 0.4586 \\
\addlinespace
\multicolumn{6}{l}{\itshape SemEval 2018 Dataset} \\
\midrule
\multirow{5}{*}{Qwen-3-8B}
  & PMP         & 0.7602 & 0.7519 & 0.7600 & 0.7541 \\
  & PMPWG       & 0.7666 & 0.7620 & 0.7730 & 0.7626 \\
  & PMPWG-FS    & 0.7653 & 0.7581 & 0.7675 & 0.7601 \\
  & PMPWL       & 0.7704 & 0.7668 & 0.7783 & 0.7669 \\
  & PMPWL-FS    & \textbf{0.7985} & \textbf{0.7912} & \textbf{0.7840} & \textbf{0.7870} \\
\addlinespace
\multirow{3}{*}{Llama-3-8B}
  & PMP         & 0.7717 & 0.7615 & 0.7612 & 0.7614 \\
  & PMPWG       & 0.7959 & 0.7875 & 0.7956 & 0.7901 \\
  & PMPWG-FS    & 0.7679 & 0.7645 & 0.7415 & 0.7480 \\
  & PMPWL       & \textbf{0.8010} & \textbf{0.7924} & \textbf{0.7999} & \textbf{0.7950} \\
  & PMPWL-FS    & 0.7742 & 0.7692 & 0.7512 & 0.7570 \\
\addlinespace
\multicolumn{6}{l}{\itshape MUStARD Dataset} \\
\midrule
\multirow{3}{*}{Qwen-3-8B}
  & PMP         & 0.6420 & 0.6426 & 0.6420 & 0.6417 \\
  & PMPWL       & 0.6580 & 0.6619 & 0.6580 & 0.6559 \\
  & PMPWL-FS    & \textbf{0.6681} & \textbf{0.6703} & \textbf{0.6681} & \textbf{0.6671} \\
\addlinespace
\multirow{3}{*}{Llama-3-8B}
  & PMP         & \textbf{0.6348} & \textbf{0.6352} & \textbf{0.6348} & \textbf{0.6345} \\
  & PMPWL       & 0.5942 & 0.5987 & 0.5942 & 0.5895 \\
  & PMPWL-FS    & 0.5594 & 0.5757 & 0.5594 & 0.5344 \\
\addlinespace
\multirow{3}{*}{Llama-3.1-8B}
  & PMP         & 0.6043 & 0.6062 & 0.6043 & 0.6027 \\
  & PMPWL       & \textbf{0.6435} & \textbf{0.6435} & \textbf{0.6435} & \textbf{0.6435} \\
  & PMPWL-FS    & 0.5986 & 0.6003 & 0.5986 & 0.5968 \\
\bottomrule
\end{tabular}
\end{adjustbox}
\\[2pt]
\footnotesize{\textit{Note:} Precision, Recall, and Macro-F1 denote \textbf{macro-averaged} metrics computed across all classes.}
\end{table*}

\section{Results}
\label{sec:results}

In this section, we present outcomes of our experiments and analyze findings for each dataset. Table \ref{tab:evaluation_all} summarizes all performance evaluation metrics.

\subsection{Indonesian Twitter Dataset}
We evaluated three variant methods on the Indonesian Twitter dataset: PMP, PMPWG, and PMPWG-FS. Using Qwen-3-8B, PMP improved over the baseline reported in the original paper, with a significant 0.1183 increase in macro-F1. This result reflects not only the advancement of modern LLMs, but also the effectiveness of PMP in adapting to culture-specific texts. However, earlier LLM attempts were inefficient as they showed high recall but very low precision, suggesting many guessed answers. In contrast, PMP demonstrated stable and meaningful predictions, as seen from the balanced precision and recall values across both labels.

Adding Google Search API grounding to PMP framework (PMPWG) produced further gains across metrics. Macro-F1 rose between PMP and PMPWG, from 0.5173 to 0.6108, with notable improvements in precision and recall. This indicates that adding external contextual information helps the model interpret culture-specific and making predictions more accurate.

Combining web grounding with few-shot demonstrations (PMPWG-FS) yielded the best performance, with a macro-F1 score of 0.6160 and an increase in accuracy of 0.0316. The inclusion of few-shot demonstrations appears to help the model capture task-specific patterns and nuances in the domain. The steady gains from PMP to PMPWG and PMPWG-FS highlight that contextual grounding and example-driven prompting both improve sarcasm detection on culturally grounded Indonesian data.

When we repeat the same experiments with Llama-3-8B, overall results are much worse than Qwen-3-8B across all methods. In many runs, between PMP and PMPWG or PMPWG-FS shows no improvement or even degrades performance. We attribute this to Llama-3’s weaker multilingual and Indonesian coverage compared to Qwen-3. The model often fails to understand input tokens and retrieved context in Indonesian, so adding web context or few-shot examples mostly adds noise.

\subsection{SemEval-2018 Task 3}
Results on the SemEval-2018 Task 3 dataset further reinforce the effectiveness and adaptability of the proposed prompting strategies even with standard English sarcasm data. All evaluated variants such as PMP, PMPWG, PMPWG-FS, PMPWL, and PMPWL-FS, produced improvements for both Qwen-3-8B and Llama-3-8B in most runs, although the gains are often small. Based on this result, we can reflect that the LLM we are using is capable enough to understand those keywords.

The baseline PMP configuration achieved a strong macro-F1 score for both models, showing a reliable result using PMP even without additional contextual augmentation. Then, the largest single improvement using Llama-3-8B comes from internal knowledge context integration (PMPWL), increasing macro-F1 from 0.7614 to 0.7950. For Qwen-3-8B, the biggest jump is seen from PMPWL with few-shot examples (PMPWL-FS), rising from 0.7541 to 0.7870. These two results show that for English datasets the word-information block combined with PMP (and optionally with few-shot) can produce meaningful gains.

By contrast, adding Google Search API grounding and or with few-shot examples (PMPWG, PMPWG-FS) helps Llama-3-8B in several cases but not all. We believe this pattern follows from model coverage. For standard English captions, the LLM often already contains the necessary background knowledge, so web retrieval adds little and can introduce noise. Differences between models reflect their respective pretraining and multilingual strengths.

Overall, SemEval-2018 confirms that structured prompting plus concise context and examples is a strong recipe for sarcasm detection in English, and that the benefit of external retrieval depends on the model and dataset.

\subsection{MUSTaRD}
Evaluation of the MUStARD dataset further supports the effectiveness of our prompting methods in standard English sarcasm data. The PMP baseline achieved a macro-F1 of 0.6417 using Qwen-3-8B. Adding contextual internal knowledge from Token Tagging and combining that with few-shot examples (PMPWL-FS) produced the best result, with macro-F1 being 0.6671. The results show a modest but consistent improvement from structured prompting plus concise context and examples. Same as Semeval, the LLM can be reflected as capable enough to understand those keywords.

For Qwen-3-8B specifically, the jump from PMP to PMPWL-FS is on the order of a few percentage points, indicating that Qwen-3 benefits from this kind of context augmentation on MUStARD. By contrast, when we use Llama-3-8B, the pattern is the opposite. Going from PMP to PMPWL or PMPWL-FS does not improve performance and can substantially degrade it, worst is from 0.6345 to 0.5344. But Llama-3.1-8B shows the reverse and even excels Qwen-3-8B. It improves best from PMP to PMPWL, going from 0.6027 to 0.6435.

We do not yet have a definitive explanation for this split. Possible causes to investigate include differences in pretraining corpora, multilingual coverage, tokenizer behavior, and instruction tuning or fine-tuning regimes between Llama-3 and Llama-3.1. These factors can change how a model interprets English tokens and how it benefits from added context. We recommend a follow-up analysis comparing token coverage on MUStARD inputs, and checking release notes or model cards for Llama-3 vs Llama-3.1 to identify training differences that could explain the divergent behavior.

\begin{table}[h]
\caption{Google Search API Retrieval Result Examples}
\centering
\renewcommand{\arraystretch}{1.3}
\begin{tabular}{|l|p{5cm}|l|}
\hline
\textbf{Words} & \textbf{Definition (translated)} & \textbf{Correct?} \\ \hline
wkwk & Wkwk is an expression of laughter that is often used in conversations on social media to show that someone is laughing, especially in the context of a funny or amusing situation. & V \\ \hline
kw & KW is an abbreviation of "Quality", which refers to counterfeit goods that imitate the original product, but the quality is not as high as the original product.& V \\ \hline
IPB & IPB is a university in Indonesia located in Bogor, with a focus on education and research in the field of agriculture and related sciences. & V\\ \hline
cie & CIE adalah singkatan dari **Commission Internationale de l'Éclairage**, sebuah organisasi internasional yang berperan dalam standarisasi dan penelitian tentang cahaya, warna, dan pencahayaan. & X\\ \hline
\end{tabular}
\label{tab:google_search_api}
\end{table}

\subsection{Google Search API}
We analyzed the actual definitions returned by our Google Search API context retrieval. Table \ref{tab:google_search_api} shows representative samples and indicates which source produced each entry. External retrieval often recovered useful meanings for slang, regional expressions, and culture-specific references common in social media, and these recovered definitions clearly helped the downstream sarcasm classifier in many cases.

However, retrieval is noisy. Manual inspection revealed incorrect or misleading definitions for some entries. For example, a Google result for the Indonesian token "cie" returned an inaccurate gloss. In Indonesian, the usage "cie" is a playful tease directed at a couple, meant to make them blush, not a literal lexical definition. Table \ref{tab:google_search_api} also reports a breakdown of sources and the proportion of manually flagged errors, illustrating that neither source is perfect and that simple filtering materially improves prompt quality. Therefore, we use not only Google Search API but also the LLM-only approach for context retrieval.

\section{Conclusion}
\label{sec:conclusion}

Our experiments show that supplying targeted context improves sarcasm detection, especially for culturally grounded data. For the Indonesian Twitter dataset, adding non-parametric knowledge using Google Search API produced the largest gains. This finding supports the claim made by the original authors of the PMP method, who noted that PMP has limitations when dealing with datasets that exhibit extreme linguistic norms. Asking the model about its own knowledge for the context retrieval step as a self-knowledge or context-awareness strategy also helps, but the improvements are smaller. Both approaches are complementary, with self-knowledge boosting awareness, and external retrieval supplying factual grounding when the model lacks background knowledge.

On standard English datasets (e.g., MUStARD, SemEval-2018), the benefits are smaller. We observed noticeable improvements from adding context and few-shot examples. These results suggest that for languages and domains well covered by the LLM’s pretraining, the model already holds much of the needed knowledge. The main value of our methods there is to make the model more self-aware and to nudge its reasoning rather than supply new facts.

Nevertheless, the use of pretrained language models is still better and more reliable for the sarcasm detection. Even after adding contextual and example-driven layers, the PMP method performs worse compared to using PLMs.

\textbf{Limitation \& Future Improvements} \hspace{2mm} As mentioned, the hardware we used to run the experiments was not capable of running models with a larger computational cost. We also did not implement every possible combination of keyword extraction and context retrieval on every dataset. In particular, some pairings were avoided where an extractor was known to perform poorly or where retrieval would provide negligible benefit. These choices were pragmatic but should be revisited with larger compute budgets.
Future improvements could also enhance both keyword extraction and especially context retrieval to perform more accurately, each with validations for a fallback pipeline, since our approach still has flaws.

\section{Acknowledgment}
We would like to express our deepest gratitude towards Bina Nusantara University for supporting us in this particular study.

\bibliographystyle{IEEEtran}
\bibliography{refs}

\newpage

\appendices

\section{Prompts}
We present the canonical PMP skeleton and then the insertion templates used to derive the hybrid variants, both in English and Indonesian. English datasets, such as Semeval and MUStARD, use English prompts, and vice versa for the Indonesian Twitter dataset.

\subsection{PMP first call ($P_{1}$)}
\noindent English prompt:\\
\noindent\fbox{%
    \parbox{\dimexpr\linewidth-2\fboxsep-2\fboxrule}{%
        You will be given a text, and will analyze the statement. Repeat back the statement to analyze. Then, analyze the following:\\
        - What does the speaker imply about the situation with their statement?\\
        - What does the speaker think about the situation?\\
        - Are what the speaker implies and what the speaker thinks saying the same thing?\\
        Finally, decide if the speaker is pretending to have a certain attitude toward the conversation.
    }%
}

\vspace{8pt}
\noindent Indonesian prompt:\\
\noindent\fbox{%
    \parbox{\dimexpr\linewidth-2\fboxsep-2\fboxrule}{%
        Kamu akan diberikan sebuah teks dan diminta untuk menganalisis pernyataan di dalamnya. Ulangi kembali pernyataan yang akan dianalisis. Kemudian, analisis hal-hal berikut:\\
        - Apa yang diimplikasikan oleh pembicara tentang situasi melalui pernyataannya?\\
        - Apa yang dipikirkan pembicara tentang situasi tersebut?\\
        - Apakah yang diimplikasikan dan yang dipikirkan pembicara menyampaikan hal yang sama?\\
        Terakhir, tentukan apakah pembicara berpura-pura memiliki sikap tertentu terhadap percakapan tersebut.
    }%
}

\subsection{PMP second call ($P_{2}$)}

\noindent English prompt:\\
\noindent\fbox{%
    \parbox{\dimexpr\linewidth-2\fboxsep-2\fboxrule}{%
        You will be given a piece of movie dialogue, a statement marked in brackets, and a preliminary analysis on the marked statement. Summarize the preliminary analysis and the given dialogue. Decide whether the statement is sarcastic or not by first analyzing the following:\\
        The Implicature – What is implied in the conversation beyond the literal meaning?\\
        The Presuppositions – What information in the conversation is taken for granted?\\
        The Intent of the Speaker – What do the speaker(s) hope to achieve with their statement and who are the speakers?\\
        The Polarity – Does the last sentence have a positive or negative tone?\\
        Pretense – Is there pretense in the speaker’s attitude?\\
        Meaning – What is the difference between the literal and implied meaning of the statement?\\
        Reflect on the preliminary analysis and what should change, then decide if the statement is sarcastic.
    }%
}

\noindent Indonesian prompt:\\
\noindent\fbox{%
    \parbox{\dimexpr\linewidth-2\fboxsep-2\fboxrule}{%
        Kamu akan diberikan sebuah pernyataan dan analisis awal terhadap pernyataan tersebut. Ringkas analisis awal tersebut. Tentukan apakah pernyataan tersebut bersifat sarkastik atau tidak dengan terlebih dahulu menganalisis hal-hal berikut:\\
        Implikatur – Apa yang tersirat dalam percakapan di luar makna literal?\\
        Presuposisi – Informasi apa dalam percakapan yang dianggap sudah diketahui?\\
        Niat pembicara – Apa yang ingin dicapai pembicara dengan pernyataannya dan siapa pembicaranya?\\
        Polaritas – Apakah kalimat terakhir bernada positif atau negatif?\\
        Kepura-puraan – Apakah ada kepura-puraan dalam sikap pembicara?\\
        Makna – Apa perbedaan antara makna literal dan makna tersirat dari pernyataan tersebut?\\
        Renungkan analisis awal dan apa yang perlu diubah, lalu tentukan apakah pernyataan tersebut bersifat sarkastik.
    }%
}

\subsection{Word-information Context Integration}
Used in PMP with word-information context, either by LLM-approach ($W_{l}$) API or Google Search API ($W_{g}$), for PMPWL and PMPWG pipelines.

\subsubsection{User Prompt}
Insert the block below in the user prompt of $P_{1}$ immediately after the input line.

\noindent English prompt:\\
\noindent\fbox{%
    \parbox{\dimexpr\linewidth-2\fboxsep-2\fboxrule}{%
        Entity facts:\\
        token1 is a ... token2 is a ... token3 is a ...
    }%
}

\vspace{8pt}
\noindent Indonesian prompt:\\
\noindent\fbox{%
    \parbox{\dimexpr\linewidth-2\fboxsep-2\fboxrule}{%
        Definisi kata-kata penting:\\
        token1 adalah ... token2 adalah ... token3 adalah ...
    }%
}

\subsubsection{System Prompt}
Also, add the block below in the system prompt of $P_{1}$, appending the last sentence.

\noindent English prompt:\\
\noindent\fbox{%
    \parbox{\dimexpr\linewidth-2\fboxsep-2\fboxrule}{%
        There are also some entity facts from the sentence that you can use. Only use them if directly relevant, do NOT invent new facts.
    }%
}

\vspace{8pt}
\noindent Indonesian prompt:\\
\noindent\fbox{%
    \parbox{\dimexpr\linewidth-2\fboxsep-2\fboxrule}{%
        Selain itu, ada disediakan beberapa fakta entitas dari kalimat yang dapat Anda gunakan. Hanya gunakan fakta tersebut jika langsung relevan, JANGAN menciptakan fakta baru.
    }%
}

\subsection{Few-shot Examples Insertion}
Used in PMPWL and PMPWG by injecting few-shot examples ($F$), with k=2, into the system prompt of $P_{2}$ after the instructions. Pipelines include PMPWL-FS and PMPWG-FS.

\noindent English prompt:\\
\noindent\fbox{%
    \parbox{\dimexpr\linewidth-2\fboxsep-2\fboxrule}{%
        Here are example reflections:\\
        Tweet: Great — the power went out during my online final. Perfect timing\\
        Implicature: The literal praise ("Great", "Perfect timing") contradicts the negative situation (power outage during an important exam); the speaker likely means the opposite.\\
        Presuppositions: It is assumed the outage disrupted the exam and caused stress.\\
        Speaker intent: To express frustration and criticize the situation indirectly, not to genuinely praise it.\\
        Polarity: Literal wording is positive, but implied polarity is negative.\\
        Pretense: There is clear pretense — the speaker is pretending to praise while actually conveying annoyance.\\
        Meaning: The literal and implied meanings diverge (literal praise vs. implied complaint), indicating irony.\\
        Final reflection: Strong contrast between wording and situation supports a sarcastic reading.\\
        Final decision: YES
        Tweet: ...
    }%
}

\vspace{8pt}
\noindent Indonesian prompt:\\
\noindent\fbox{%
    \parbox{\dimexpr\linewidth-2\fboxsep-2\fboxrule}{%
        Tweet: Bagus banget, listrik mati pas lagi final online. Sumpah rejeki beneran\\
        Implikatur: Kalimat tampak memuji ("Bagus banget") tetapi konteks (listrik mati saat final) jelas negatif; pembicara menyindir situasi.\\
        Presuposisi: Diasumsikan listrik mati dan menimbulkan masalah pada ujian/online.\\
        Niat pembicara: Mengungkapkan kekesalan dengan ironi, bukan benar-benar memuji.\\
        Polaritas: Literal positif, implisit negatif.\\
        Kepura-puraan: Ada pretense — pura-pura menyatakan kejelekan sebagai "bagus".\\
        Makna: Perbedaan jelas antara makna literal (pujian) dan tersirat (keluhan).\\
        Refleksi akhir: Kontras kata vs konteks kuat; indikator sarkastik jelas.\\
        Keputusan akhir: YES\\
        Tweet: ...
    }%
}

\subsection{LLM-Based Keyword Extraction}
Used in PMP with LLM-based keyword extraction, where pipelines include PMPWG and PMPWG-FS.

\subsubsection{Keyword Choosing}
Query the block below as the system prompt to the LLM with the user prompt being the target input text.

\noindent English prompt:\\
\noindent\fbox{%
    \parbox{\dimexpr\linewidth-2\fboxsep-2\fboxrule}{%
        You will be given a text containing a list of unknown words. Your task is to separate the words into comma-separated values (CSV). If there are no unknown words, answer with 'NO UNKNOWN'\\
        example output:\\
        first,second,third
    }%
}

\vspace{8pt}
\noindent Indonesian prompt:\\
\noindent\fbox{%
    \parbox{\dimexpr\linewidth-2\fboxsep-2\fboxrule}{%
        Anda akan diberikan teks berisi penjelasan kata-kata yang tidak dimengerti. Tugas anda adalah memisahkan kata-kata tersebut menjadi daftar yang dipisahkan koma (CSV). Kalau tidak ada kata-kata yang tidak dimengerti, jawab dengan 'NO UNKNOWN'\\
        contoh output:\\
        pertama,kedua,ketiga
    }%
}
\\
\subsubsection{Choosement Cleaning}
After the LLM-based keyword extraction, query the LLM again with this block below as the system prompt for cleaning, with the user prompt being the output from the previous call, resulting in a clean list of unknown keywords according to the LLM.

\noindent English prompt:\\
\noindent\fbox{%
    \parbox{\dimexpr\linewidth-2\fboxsep-2\fboxrule}{%
        You will be given a text. Your task is to identify words from the text that you do not understand.
    }%
}

\vspace{8pt}
\noindent Indonesian prompt:\\
\noindent\fbox{%
    \parbox{\dimexpr\linewidth-2\fboxsep-2\fboxrule}{%
        Anda akan diberikan sebuah teks. Tugas anda adalah menyebutkan kata-kata yang anda tidak mengerti dari teks tersebut.
    }%
}

\section{Data Availability}
The datasets used in this study are publicly available as follows:
\begin{itemize}
    \item \href{https://github.com/Cyvhee/SemEval2018-Task3}{SemEval-2018 Task 3} \cite{semeval}
    \item \href{https://github.com/soujanyaporia/MUStARD}{MUStARD} \cite{mustard}
    \item \href{https://huggingface.co/datasets/w11wo/twitter_indonesia_sarcastic}{Twitter Indonesia Sarcastic} \cite{suhartono2024idsarcasm}
\end{itemize}
For further details regarding the experiments, please contact Michael Iskandardinata via email.

\section{Author contributions}
We follow the CRediT taxonomy to state individual contributions, with roles in parentheses.
\begin{itemize}
    \item Michael Iskandardinata (first author): Conceptualization (equal); Methodology (equal); Software (equal); Validation (support); Resources (lead); Investigation (lead); Data Curation (support); Writing - Review \& Editing (equal); Visualization (lead);
    \item William Christian (co-author): Conceptualization (equal); Methodology (equal); Software (equal); Validation (lead); Resources (support); Investigation (support); Data Curation (lead); Writing - Review \& Editing (equal); Visualization (support);
    \item Derwin Suhartono (supervisor): Conceptualization (equal); Writing - Review \& Editing (equal); Supervision.
\end{itemize}

Notes: the primary inference designs and prompt engineering were done by Michael Iskandardinata. Both authors collaborated on Token Tagging and LLM retrieval code; William Christian implemented the Google Search API retrieval pipeline with LLM-cleaning. All authors reviewed and approved the final manuscript.

\end{document}